\documentclass[a4paper,10pt]{article}
\usepackage{amsmath}
\usepackage{amssymb}
\usepackage[dvipdfmx]{graphicx}
\usepackage{ifthen}
\usepackage{verbatim}
\usepackage{color}
\usepackage{slashbox}
\usepackage{url}

\begin{document}

\bigskip
\bigskip
\begin{center}
{\Large \textbf{
	Locality-Sensitive Hashing with Margin Based Feature Selection
}}
\end{center}

\vspace{10mm}
\begin{center}
Makiko Konoshima,
\footnote{E-mail: \texttt{makiko@jp.fujitsu.com}}
Yui Noma
\\
\bigskip
{\small 
\textit{
Software Systems Laboratories, FUJITSU LABORATORIES LTD.  1-1,\\
Kamikodanaka 4-chome, Nakahara-ku Kawasaki, 211-8588 Japan.\\}}
\end{center}

\vspace{10mm}
\begin{abstract}


We propose a learning method with feature selection for
 Locality-Sensitive Hashing. 
Locality-Sensitive Hashing converts feature vectors into bit arrays. 
These bit arrays can be used to perform similarity searches and personal authentication. 
The proposed method uses bit arrays longer than those used
 in the end for similarity and other searches and
 by learning selects the bits that will be used. 
We demonstrated this method can effectively perform optimization for cases
 such as fingerprint images with a large number of labels
 and extremely few data that share the same labels,
 as well as verifying that it is also effective for natural images,
 handwritten digits, and speech features.

{\bf Keyword:} Locality-sensitive hashing, Feature selection, High-dimensional data

\end{abstract}

\vspace{10mm}

\section{Introduction}

\begin{figure*}[tb]
	\begin{center}
	\includegraphics[scale=0.3]{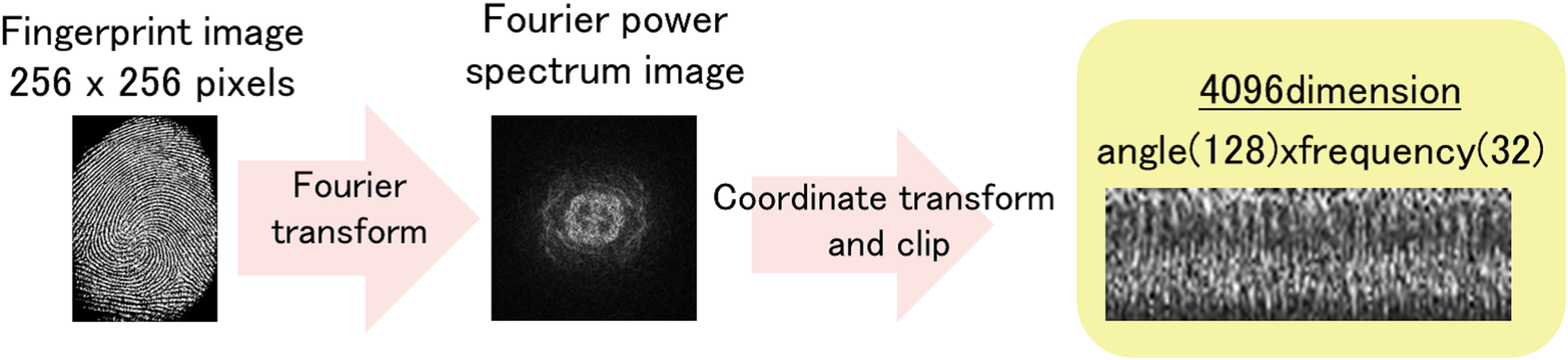}
	\end{center}
	\caption{Flow chart of fingerprint image feature generation}
	\label{FingerprintPower}

	\begin{center}
	\includegraphics[scale=0.3]{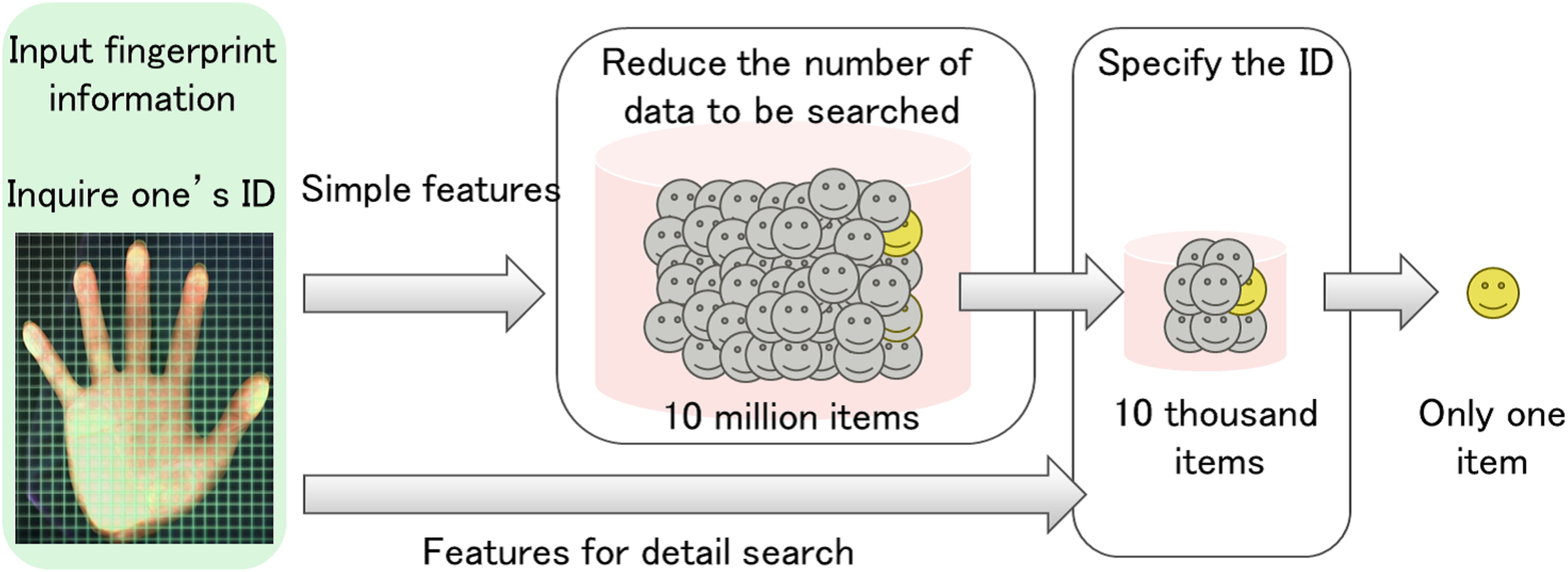}
	\end{center}
	\caption{Flow chart of fingerprint image identification using a refined search}
	\label{idByFinger}
\end{figure*}

Recently, biometric authentication techniques have become widely
 used to prevent information leaks from companies and spoofing fraud
 at financial institutions~\cite{Seitai_JRP}. 
Because of their high authentication performance,
 they have a wide variety of applications, domestically and overseas,
 such as identity verification at ATMs,
 and PC and room access controls at companies. 
The 1:N identification service can identify an individual by only using biological
 information, which is one of the advantages of biometric authentication. 
This technique is expected to be used widely in the near future because of its convenience. 
A biometric sensor obtains different biological features every time it works. 
Therefore, it is necessary for 1:N identification to calculate similarities
 using all data in the database and extract the data most similar to
 those that are collected from the person to be identified. 
This causes a problem because this process involves all data of N people resulting
 in a huge amount of calculation operations. 
So, it is important to make in advance a short list of a few hundredths or a few thousandths
 of the total data searched. 
To keep comparison operations fast and accurate enough at the same time,
 simple features are used for refined searches,
 and intricate features are used for detailed searches. 
In practice, a fingerprint authentication method uses fingerprint image power spectra
 as features~\cite{Finger_power}(Fig.~\ref{FingerprintPower}),
 and refined searches using this method have
 achieved great success~\cite{Seitai_FujitsuPress} (Fig.~\ref{idByFinger}). 
Thus, high-speed similarity searches in a high-dimensional
 feature space are an important research subject. 
Search methods for high-speed huge-data searches include KD-tree~\cite{KDTree} 
and iDistance~\cite{iDistance}. 
However, these methods have not solved the problem of long processing times
 when calculating Euclidean distances in high-dimensional data searches. 
Locality-Sensitive Hashing~\cite{LSH_IndykMotwani} has been proposed to solve these problems and
 is attracting attention as a technology capable of high-speed similarity searches
 for high-dimensional data.

Locality-Sensitive Hashing converts a huge number of high-dimensional feature vectors
 into bit arrays to carry out high-speed similarity calculations using the Hamming distance. 
A typical method for Locality-Sensitive Hashing is Random Projection
 (this method is hereinafter called LSH in this paper)~\cite{LSH_RandomProjection}. 
This method partitions a feature space with hyperplanes. 
Each point in a feature space is assigned bits according to the signs of
 the inner products with respect to the normal vectors of the hyperplanes. 
Therefore, the number of the hyperplanes is the same as the length of the bit arrays. 
Reference~\cite{LSH_RandomProjection} treats a feature space
 as a vector space and discusses only hyperplanes
 that cross the origin. 
It is known that, in this method, as the number of bits increases,
 the correlation between Hamming distances and angles becomes stronger. 
For this reason, increasing bit numbers improves search performance for data sets
 in which angles between features correspond to dissimilarities. 
However, these angles may not represent dissimilarities correctly
 when identifier labels are assigned to data. 
In this case, determining hyperplanes by learning can improve the search accuracy.


Several learning methods for hyperplanes have been proposed. 
However, learning methods by optimization, such as Minimal Loss Hashing~\cite{MLH},
 need to assume differentiability of object functions. 
Additionally, learning methods by optimization have the following problems in general:
 increasing the number of bits to improve search accuracy leads to
 a large solution space resulting in calculation that take too much time;
 and too much attention to local optimization hinders the goal of obtaining a global optimum solution.


This paper proposes a hyperplane selection method that uses feature selection. 
This method avoids assuming differentiability of object functions,
 and, for large numbers of bits, can perform optimization without searching vast solution spaces.


To compare the proposed method with related conventional methods,
 search performances were calculated for the following data sets in this paper:
 MINIST handwritten character database~\cite{MNIST},
 LabelMe~\cite{LabelMe},
 features resulting from Fourier transform of fingerprint images~\cite{Finger_power},
 and Mel Frequency Cepstral Coefficient (MFCC) features of speech~\cite{SpeechMFCC}.

\section{Background and related work}
\label{sec_relatedWork}


This chapter describes 
 to Locality-Sensitive Hashing using random projection,
 which the proposed method is based upon,
 and conventional methods related to it.

\subsection{Random projection}

Let $V$ be an $N$-dimensional vector space that has high-dimensional feature vectors. 
For simplicity, consider more than one hyperplanes that cross the origin of $V$.
A hyperplane that crosses the origin can be described in terms of a normal vector.


LSH generates $B$ normal vectors randomly. 
The $N$-dimensional feature vector
 $\vec{x}$
 is converted into a bit array
 so that the data are assigned 1
 if the signs of the inner products with respect to the normal vectors are positive,
 and 0 if otherwise. 
This conversion can be expressed as follows. 
Let $W$ be a $B \times N$ matrix, assume its row vectors are normal vectors,
 and bit array b is given by:
\begin{eqnarray}
	b(\vec{x}) = thr \left( W \cdot \vec{x} \right) ,
\end{eqnarray}
where $thr$ is a function that returns a bit array,
 and its $i$-th component is 1 if the $i$-th argument vector component is positive,
 and 0 if otherwise.


We call this method LSH. 
Learning methods based on LSH aim to determine matrix $W$.

\subsection{Learning with optimization}


Minimal Loss Hashing (MLH) is a supervised learning of hyperplanes~\cite{MLH}. 
MLH conducts learning aiming to minimize a discontinuous function called
 the empirical loss function that has $W$ as its argument. 
The empirical loss function has a large value when data pairs
 with the same labels have large Hamming distances,
 and data pairs of different labels have small Hamming distances. 
Since the empirical loss function is a discontinuous function,
 optimization with the gradient method cannot be applied. 
For this reason, in MLH one considers a differentiable upper bound function of
 the empirical loss function and minimizes
 the upper bound function by the stochastic gradient method.


Principal Component Analysis Hashing (PCAH)~\cite{PCAH}
 is a unsupervised learning to determine hyperplanes. 
PCAH analyzes principal components of learning data to make principal
 component vectors to be normal vectors of hyperplanes. 
A disadvantage of this method is that it cannot treat bit numbers larger than
 the dimension of the feature space.

\subsection{Learning with feature selection for each class}


Methods that prepare a number of hyperplanes and select hyperplanes
 to be used for each query type are proposed in References~\cite{FS1} and~\cite{FS2}. 
These methods need to discriminate query types and need to learn
 the selection of hyperplanes to be used for each query type.

\subsection{Other hashing schemes}


There are hashing schemes called Kernelized LSH (KLSH)~\cite{KLSH}
 as an extended LSH that uses Kernel functions,
 and Spectral Hashing (SH)~\cite{SH} that is a hashing scheme
 with eigenfunctions determined by data distributions only.


Since KLSH uses kernel functions,
 it has to process a large amount of calculations in general,
 which takes a long time to carry out the conversion of a large number of bits into bit arrays.


SH makes trigonometric functions in the data space
 and uses the signs of their function values to carry out the conversion into bit arrays. 
It uses high-frequency functions for a large number of bits. 
So, for high-frequencies, many data pairs with large L2 distances
 from each other are mapped to the same bit values. 
Also, in the case that feature vectors have a slight amount of noise,
 the high frequency trigonometric functions may cause mapping these vectors
 to bit values far different from the original values. 
Thus, SH performance is significantly poor in conditions with a large number of bits.

\section{Locality-Sensitive Hashing with margin based feature selection}


We propose a supervised learning method of hyperplanes using feature selection
 (hereinafter called S-LSH). 
This method does not need differentiability of object functions,
 and avoids searching vast solution spaces. 
As can be easily seen, feature selection using the proposed method
 can be applied to other hashing schemes (such as Spectral hashing)
 than hashing using hyperplanes.


We referred to the Interactive Search Margin Based Algorithm~\cite{MarginBasedFS}
 for basic information of feature selection. 
This chapter describes below how we applied the method of Reference~\cite{MarginBasedFS}
 to hyperplane selection. 
For details, refer to the literature.


An outline of the learning algorithm is as follows. 
Generate hyperplanes randomly in number of $\tilde{B}$
 sufficiently larger than the target bit number $B$. 
Assign parameters called the degree of importance to these hyperplanes. 
Consider Hamming distances with degrees of importance and update
 the degrees of importance so that importance-attached Hamming distances of
 data pairs of identical labels become smaller,
 and those of different labels become larger. 
After repeating learning a certain number of times,
 select $B$ hyperplanes in the order of importance.


The learning method is explained in detail as follows. 
Assign degrees of importance $\left\{ \omega_i \right\}_{1\leq i \leq \tilde{B}}$
 to $\tilde{B}$ hyperplanes. 
Set all initial values to 1.
Feature vectors can be converted into bit arrays
 in the same manner as in LSH.
However, this method gives bit arrays of a length of $\tilde{B}$,
 which is generally longer than $B$ that will be obtained in the end.
For bit array $z$, weighted Hamming distance $||z||_\omega$
 is defined by the following formula:
\begin{eqnarray}
	||z||_\omega := \sqrt{ \sum_{i=1}^{\tilde{B}} \omega_i^2 z_i^2 }.
\end{eqnarray}
Let $x$ be a bit array for a data point,
 and then let $nearhit(x)$ be the data with the smallest weighted Hamming distance
 among those having a label identical to that of $x$,
 and let $nearmiss(x)$ be the smallest weighted Hamming distance among
 those having a label different from that of $x$. 
The learning process aims to maximize the value of the following formula:
\begin{eqnarray}
	&&e(\omega) := \sum_{x \in S} \theta_{S\setminus \{x\} } (x), \\
	&&\theta_P(x) := \nonumber\\
		&&\frac{1}{2}
		\left( ||x- nearmiss(x)||_\omega - ||x-nearhit(x)||_\omega \right) ,
	\label{eq_eOmega}
\end{eqnarray}
where $S$ is the set of learning data. 
By this method, learning hyperplanes can be reduced to an optimization problem
 in the $\tilde{B}$-dimensional space.


Theoretically, maximizing $e(\omega)$ means the following. 
In general, when hyperplanes partition two data points,
 Hamming distance between the two points increases. 
Different labels are given to $x$ and $nearmiss(x)$ and
 they should have a large Hamming distance and be partitioned by hyperplanes preferably. 
In other words, the degrees of importance of hyperplanes
 that partition $x$ and $nearmiss(x)$ should be large. 
Also, $x$ and $nearhit(x)$ that have the same label should have
 a small Hamming distance and not be partitioned by hyperplanes preferably. 
Therefore, the degrees of importance of hyperplanes
 that partition $x$ and $nearhit(x)$ should be small. 
The larger the degrees of hyperplanes that partition $x$ and $nearmiss(x)$,
 the first term of equation~(\ref{eq_eOmega}) returns a larger value. 
The smaller the degrees of hyperplanes that partition $x$ and $nearhit(x)$,
 the second term of equation~(\ref{eq_eOmega}) including the sign returns a larger value. 
Consequently, maximizing $e(\omega)$ means giving the degrees of desirable hyperplanes.


Use the stochastic gradient method to maximize $e(\omega)$. 
Update the degree of importance $\omega$ for point $x$ randomly selected from the learning data. 
Use the following formula to update the degree of importance $\omega$:
\begin{eqnarray}
	&&\delta \omega_i = \nonumber\\
	&&\frac{1}{2}\sum_{i=1}^{\tilde{B}}
	\left(
		\frac{(x_i - nearmiss(x)_i)^2}{||x-nearmiss(x)||_\omega}
		- \frac{(x_i - nearhit(x)_i)^2}{||x-nearhit(x)||_\omega}
	\right) \omega_i. \nonumber\\
\end{eqnarray}
After repeatedly updating a specified number of times,
 select $B$ hyperplanes in the order of absolute value of degree of importance. 
The number of updates is called the number of learning.
For the reason mentioned earlier,
 this process selects hyperplanes that do not partition data pairs
 with the same labels and partition data pairs of different labels.


As can be seen from the experiments below,
 it is empirically clear that carrying out
 learning 10,000 times ensures sufficient performance.

\section{Experiments}


To evaluate the performance of the proposed method (S-LSH),
 we carried out experiments to compare the method with
 LSH, SH, and MLHamong the conventional methods mentioned in section~\ref{sec_relatedWork}
We also compared the searching with Euclidean distances (L2), which is the traditional method. 
The following data are used:
 GIST features of LabelMe~\cite{GIST},
 MNIST handwritten character database~\cite{MNIST},
 MFCC features of speech~\cite{SpeechMFCC},
 and features resulting from Fourier transform of fingerprint images~\cite{Finger_power}. 
As a performance measurement, we calculated precision and recall curves for each method. 
Precision and recall are defined by the following formulas:
\begin{eqnarray}
	Precision &:=& \frac{
				\begin{array}{l}
					\mbox{Number of data of the same label} \\
					\mbox{as the query acquired by search}
				\end{array}}
			{\mbox{Number of data acquired by search}} ,\\
	Recall &:=& \frac{
				\begin{array}{l}
					\mbox{Number of data of the same label}\\
					\mbox{as the query acquired by search}
				\end{array}}
			{\begin{array}{l}
				\mbox{Number of data of the same label}\\
				\mbox{as the query of data searched}
			\end{array}} .
\end{eqnarray}
Also, a search was carried out in a database,
 and search results acquired from the data searched were sorted
 in ascending order of distance with respect to a query. 
The rate of search results is defined as the acquisition:
\begin{eqnarray}
	Acquisition := \frac{\mbox{Number of data acquired by search}}
					{\mbox{Total number of data searched}}.
\end{eqnarray}


The experiment procedure was as follows. 
The data was affine-transformed so that the mean of the respective components
 is 0 and the standard deviation is 1. 
Principal component analysis (PCA) was performed on the transformed data. 
The data was projected to the subspace where the cumulative contribution ratio exceeds 80\%. 
By using the projected data, the hyperplanes were learned. 
As the labeling varied depending on data sets,
 it is described separately in each experiment explanation below. 
With searches in mind, we divided the data into the following three types:
 data to be used for learning (hereinafter called learning data),
 data registered in the database (hereinafter called test data),
 and data as queries for searches (hereinafter called query data). 
These three types of data sets were made mutually disjoint.
The numbers of bits were 32, 64, 128, 256, 512, and 1,024.
For S-LSH, $\tilde{B} = 10,000$,
 and the number of learning was 10,000. MLH carried out learning $10^7$ times.

\subsection{Experiments on LabelMe}


We used a LabelMe data set of 512-dimensional Gist features~\cite{GIST}
 extracted from image data, which is described in Reference~\cite{512dimGISTdata},
 to carry out the experiments. 
The labeling was done to the given non-similarity matrix of data
 so that the top 50 pairs of each line have the same label as the line. 
Of the 22,000 data in total, 11,000 were used for learning, 5,500 for query,
 and 5,500 for test data. PCA reduced the dimension to 20.

\begin{figure*}[tb]
	\begin{center}
	\includegraphics[scale=0.5]{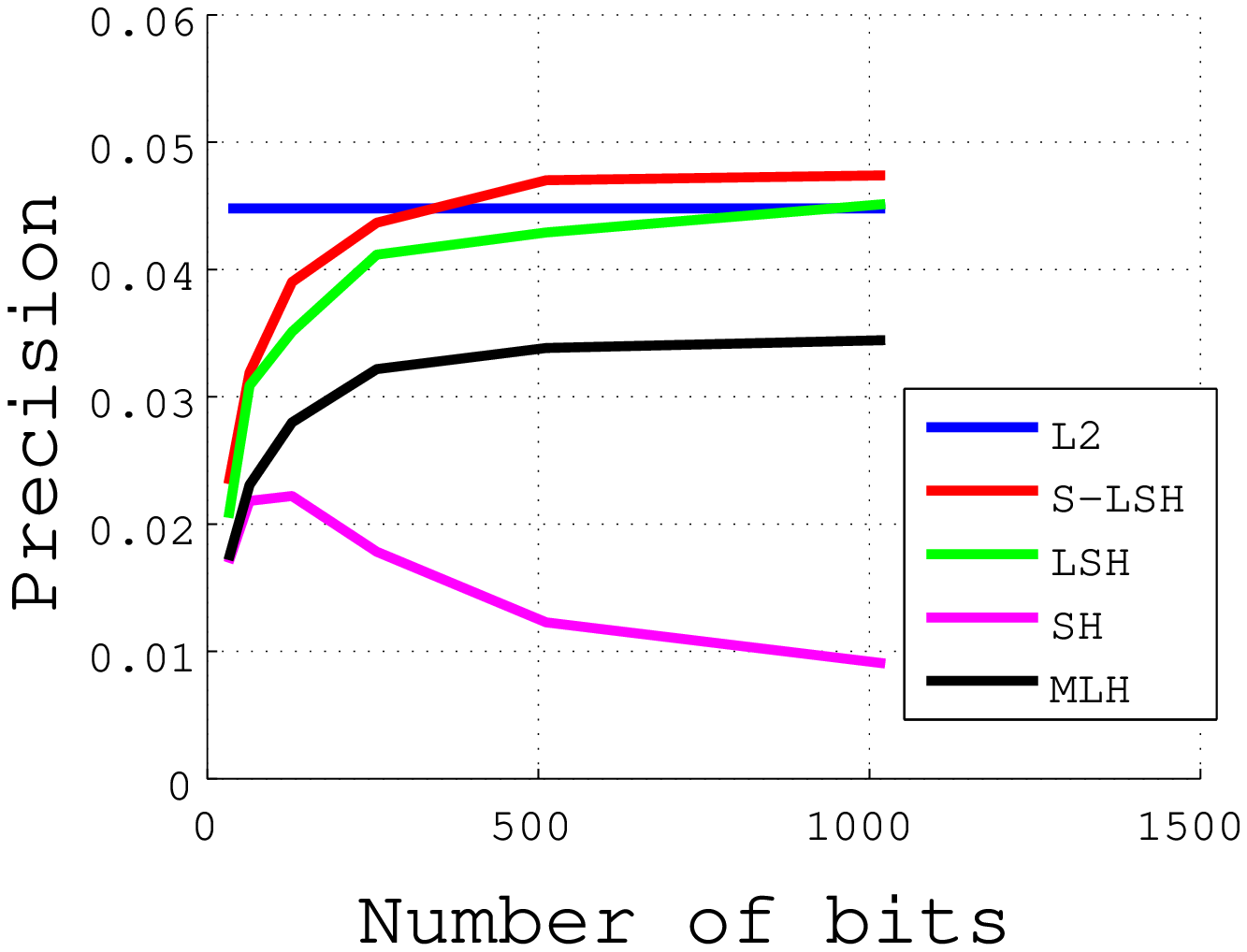}
	\includegraphics[scale=0.5]{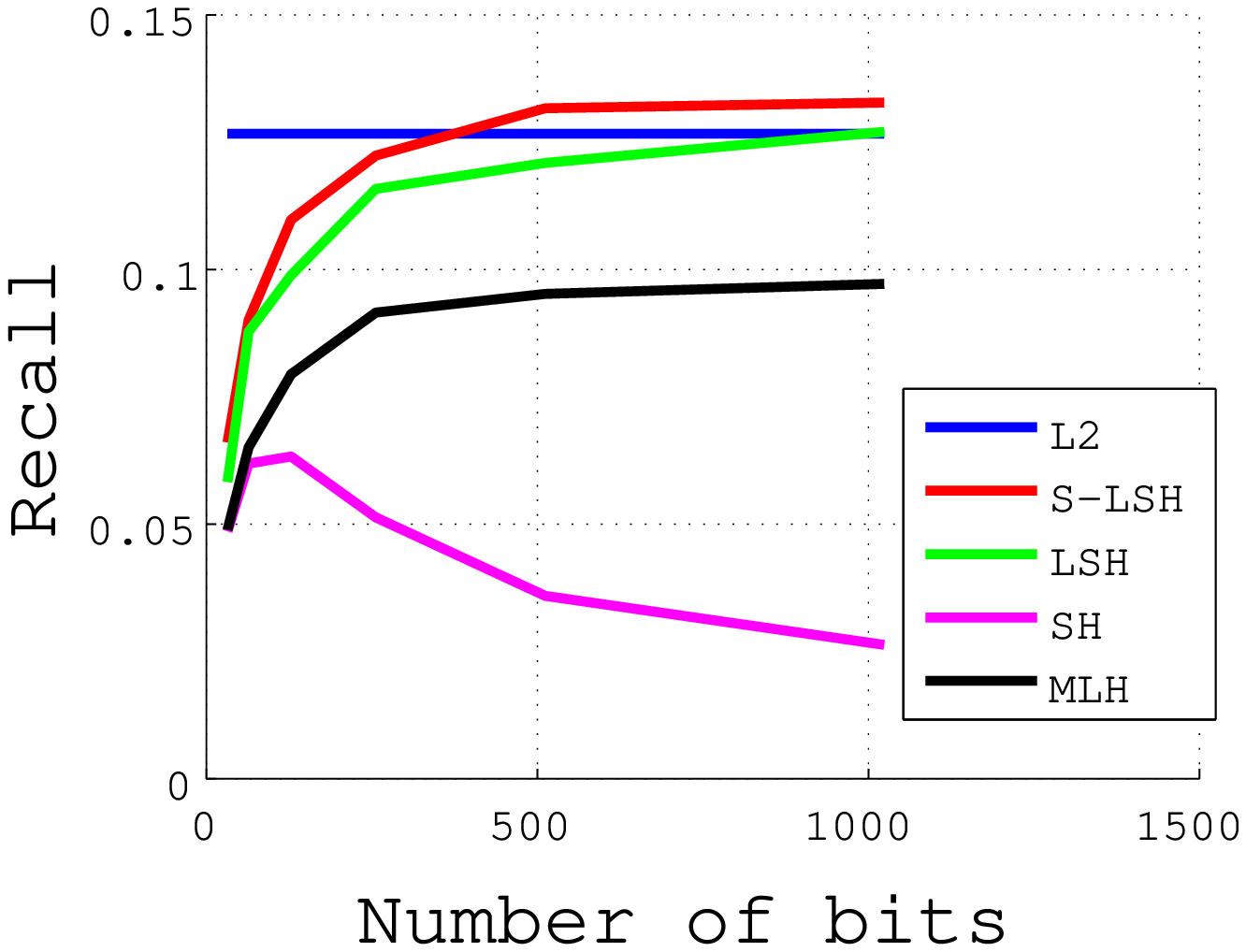}
	\end{center}
	\caption{LabelMe: Curves of precision (left) and recall (right)
		versus the number of bits for L2, S-LSH, LSH, SH, and MLH.
		}
	\label{fig_LabelMePR}
\end{figure*}

Fig.~\ref{fig_LabelMePR} shows graphs of dependency of the precision and the recall
 on the number of bits with the acquisition fixed at 0.01. 
S-LSH is observed performing well. 
MLH shows no learning performance improvements,
 and SH performs poorer as the number of bits increase.

\subsection{Experiments on MNIST}


MNIST is a set of 8-bit grayscale images of handwritten digits,
 0 to 9, consisting of $28 \times 28$ pixels per digit,
 which are individually assigned labels of 0 to 9. 
The numbers of data used were:
 60,000 for learning data,
 5,000 for query data,
 and 5,000 for test data.


We used these image data as features to study precision and recall.
PCA reduced the dimension to 149.

\begin{figure*}[tb]
	\begin{center}
	\includegraphics[scale=0.5]{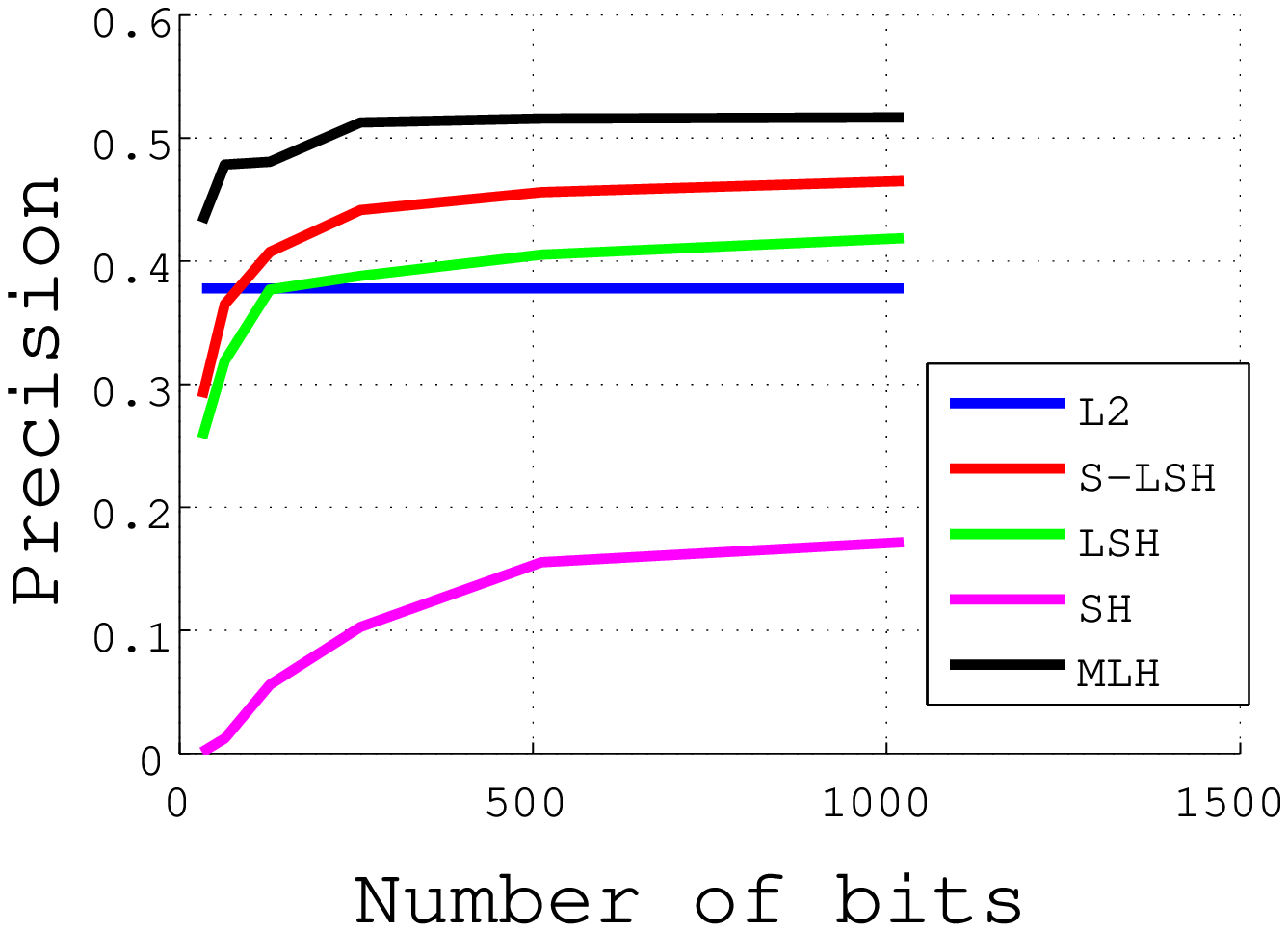}
	\includegraphics[scale=0.5]{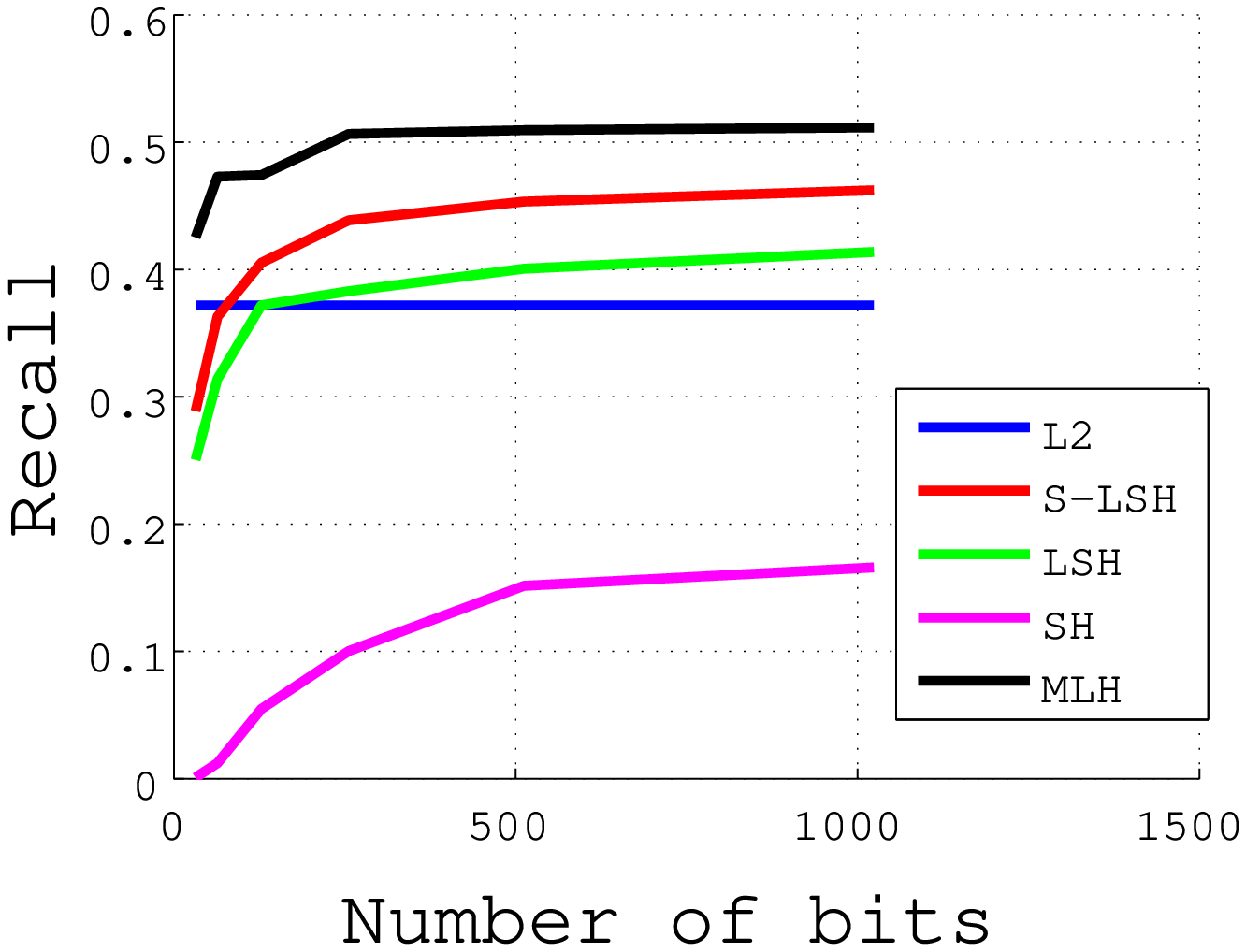}
	\end{center}
	\caption{MNIST: Curves of precision (left) and recall (right)
				versus the number of bits for L2, S-LSH, LSH, SH, and MLH.
			}
	\label{fig_MNISTPR}
\end{figure*}

Since MNIST has ten types of labels,
 Fig.~\ref{fig_MNISTPR} shows graphs of dependency of the precision and recall on
 the number of bits with the acquisition fixed at 0.1. 
Although inferior to MLH, S-LSH performs better than LSH. 
SH performs poorer as the number of bits increases.

\subsection{Experiments on speech features}


For the experiment, we used Internet-available recorded data
 of a three hour long local government council meeting~\cite{KawasakiCouncil}. 
Because speech data is temporally continuous,
 we used 200-dimensional MFCC data obtained by using overlapping window functions as features.
For the queries, we used speech sounds obtained separately.


For supervised learning of speech, speech content (text) is usually used as labels.
For this study, however, we assumed a collection of data with similar features,
 and those features with the top 0.1\% shortest Euclidean distances from the queries
 were regarded in the same class for conducting learning and evaluation. 
For learning, 192,875 MFCC data obtained from 378 speech data were used. 
The number of data is 1,815 for query data, 192,683 for data searched. 
PCA reduced the dimension to 30.

\begin{figure*}[tb]
	\begin{center}
	\includegraphics[scale=0.5]{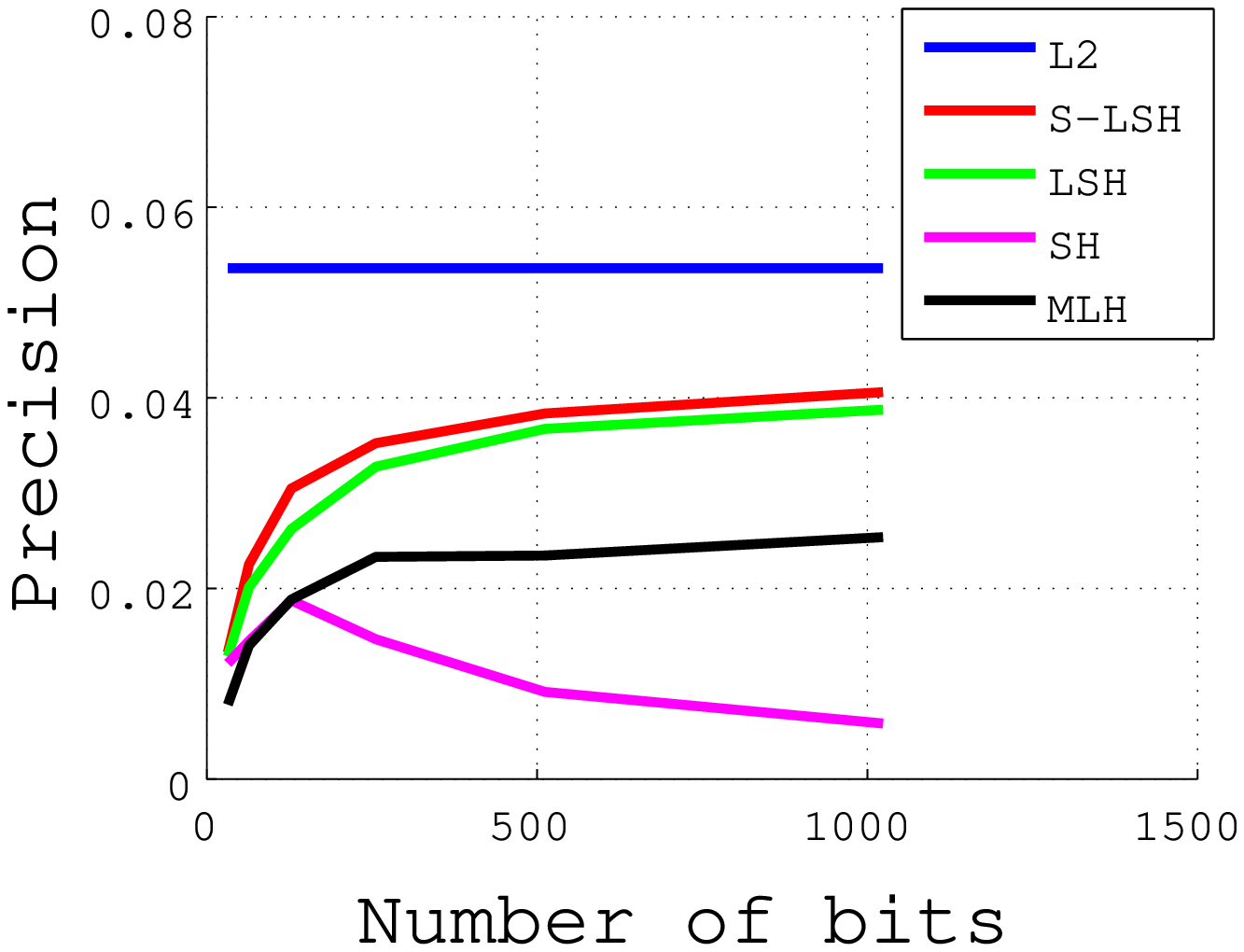}
	\includegraphics[scale=0.5]{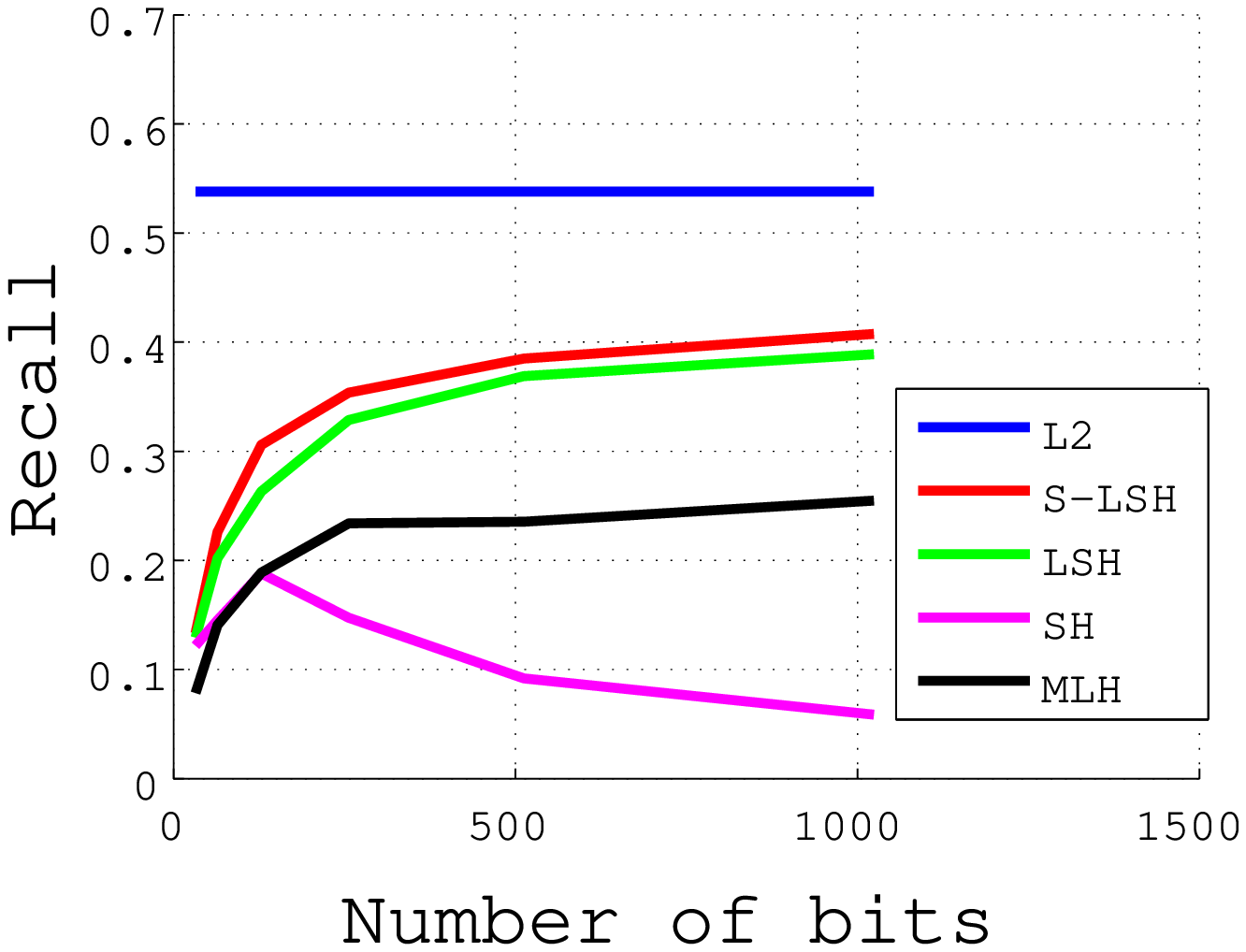}
	\end{center}
	\caption{Speech features: Curves of precision (left) and recall (right) versus number of bits
			for L2, S-LSH, LSH, SH, and MLH.
			}
	\label{fig_SpeechPR}
\end{figure*}

Fig.~\ref{fig_SpeechPR} shows graphs of dependency of the precision and recall on the number of bits
 with the acquisition fixed at 0.01. 
S-LSH performs better.
MLH shows no learning performance improvements,
 and SH performs poorer as the number of bits increases.

\subsection{Experiments on fingerprint images}


Since biometric authentication using fingerprint data deals with a huge number of search objects,
 it needs to use a refined search used for 1:N identification. 
For refined searches, a search result has to include an ID that corresponds to the query.
We carried out the experiments with this premise in mind. 
As shown below, the error rate is defined as the probability that
 the data obtained by the search do not include data
 that have the same labels as the queries do.
\begin{eqnarray}
	&&Error\: rate := \nonumber \\
		&&	\frac{
				\begin{array}{l}
					\mbox{Number of query data whose labels are}\\
					\mbox{not assigned to the search result data}
				\end{array}}
			{\mbox{Total number of query data}}.
				\nonumber\\
\end{eqnarray}
The error rate is a factor that indicates the accuracy of the refined search. 
The smaller this value, the better the accuracy is.


We collected fingerprint images on our own with a fingerprint reader. 
The features of fingerprint images were 4,096-dimensional floating point vector data
 which were made by clipping the power spectrum of the image data. 
The following describes how fingerprint image data was collected. 
Twelve image data was collected for each of the right and left second,
 third, and fourth fingers of 1,032 people. 
Fingerprint images collected that had a poor image quality were not used
 and the same label was shared by up to 12 data. 
Labels were determined for each finger of an individual at the time of collection, 
 which means that labels can be automatically assigned. 
Because the biological features of the respective fingers,
 as well as the right and left hands, are mutually independent,
 even for one person, the number of labels is 6,192.
PCA reduced the dimension to 276.


The learning used the data of approximately 25\% of the total number of people.
The experiments used 9,906 data for learning, 12,138 for test data, and 19,932 queries.

\begin{figure*}[tb]
	\begin{center}
	\includegraphics[scale=0.5]{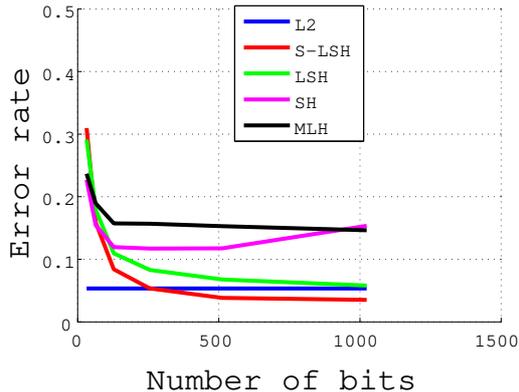}
	\end{center}
	\caption{Fingerprint features: Error rate versus number of bits for L2, S-LSH, LSH, SH, and MLH.}
	\label{fig_FingerError}
\end{figure*}

Fig.~\ref{fig_FingerError} shows graphs of dependency of the error rate on the number of bits
 with the acquisition fixed at 0.1. 
S-LSH shows improvements in learning performance,
 and has error rates lower than those of LSH and L2 as the number of bits increases. 
MLH and SH are poor in learning performance in the range of large bit numbers.

\section{Processing time of each method}


The processing times for learning of S-LSH, LSH, MLH, and SH are listed in Table~\ref{table_ProcessTime}.
The process of each method was written in the C++ programming language. 
The CPU was an Intel Xeon X5680 3.33 GHz, and its one core alone did the job (single thread). 
After reducing the dimension to a level where the cumulative contribution ratio
 reaches 80\%, the computer calculated each learning time with the bit number = 1,024. 
Other parameters were the same as those used in the experiments mentioned earlier.


Because the learning process of S-LSH examines distances for all the learning data
 that have been converted to 10,000 bits ($=\tilde{B}$),
 the learning time depends on the number of learning data,
 and linearly depends on the number of learning and $\tilde{B}$.
Because the learning process of MLH selects sample data pairs,
 the learning time linearly depends on the number of learning,
 the bit number $B$, and the dimension of feature space.

\begin{table*}[tb]
	\begin{center}
	\caption{Processing time of each method (s)}
	\begin{tabular}{c|r|r|r|r}
	\hline
	\hline
	\backslashbox{Method}{Data set} & LabelMe & MNIST & Speech & Fingerprint \\
	\hline
	S-LSH & 8698.7 & 54814.4 & 55779.6 & 9131.7 \\
	LSH & 0.0 & 0.0 & 0.0 & 0.0 \\
	SH & 0.0 & 0.1 & 0.0 & 0.1 \\
	MLH & 5533.9 & 31101.4 & 5924.55 & 53917.8 \\
	\hline
	\end{tabular}
	\label{table_ProcessTime}
	\end{center}
\end{table*}

\section{Discussion}


The proposed method S-LSH performs better than LSH for all the data sets tested. 
Table~\ref{table_RankingForMethod} shows the performance rank order numbers of
 the methods tested in this study. 
The ranking in the table is for the precision or error rate when a bit number
 of 1,024 and the acquisition mentioned in each experiment are used.

\begin{table*}[tb]
	\begin{center}
	\caption{Ranking of methods}
	\begin{tabular}{c|r|r|r|r}
	\hline
	\hline
	\backslashbox{Method}{Data set} & LabelMe & MNIST & Speech & Fingerprint \\
	\hline
	S-LSH & 1 & 2 & 1 & 1 \\
	LSH   & 2 & 3 & 2 & 2 \\
	SH    & 4 & 4 & 4 & 4 \\
	MLH   & 3 & 1 & 3 & 3 \\
	\hline
	\end{tabular}
	\label{table_RankingForMethod}
	\end{center}
	\begin{center}
	\caption{Approximate number of labels and label cardinality.}
	\begin{tabular}{c|r|r|r|r}
	\hline
	\hline
	Data set & LabelMe & MNIST & Speech & Fingerprint \\
	\hline
	Number of learning data
		& 11000 & 60000 & 192875 & 9906 \\
	Approximate number of labels
		& 300 & 10 & 2000 & 1300 \\
	Approximate cardinality of subsets for each label
		& 40 & 6000 & 100 & 7 \\
	\hline
	\end{tabular}
	\label{table_LabelNum}
	\end{center}
\end{table*}


Only MLH performed better than S-LSH for MNIST data only,
 among the data sets tested in this study. 
MLH is poorer than LSH in performance for data sets other than MNIST data.
It is in the number of labels and the concentration of data subsets
 with the same labels (hereinafter called subsets for each label)
 where MNIST data differs most from other data sets. 
However, because data sets in which a label is uniquely assigned to each data
 do not account for all data sets, the number of labels cannot be defined
 in a straightforward way. 
So, the following idea helps. 
The average number of data with the same label in the learning data is
 defined as the “approximate cardinality of subsets for each label.”
And the number of learning data divided by the approximate cardinality
 of subsets for each label is defined as the “approximate number of labels.”
Table 3 shows the approximate numbers of labels and approximate cardinality
 of subsets for each label and each type of data set.
As can be seen from Tables 2 and 3,
 MLH has poor learning performance for most data sets except for those
 that have a small number of labels and large cardinality of subsets for each label.


From the discussion above, the proposed learning method is effective for many types of data sets.
It is probably most effective among others for data sets
 with a large number of labels
 and small cardinality of subsets for each label
 that conventional methods cannot learn.

\section{Conclusion}


This paper has proposed a method that selects data from generated hyperplanes
 that outnumber the target hyperplanes for data conversion of feature vectors
 into bit arrays using hyperplanes. 
We demonstrated that this proposed method is highly effective even for data sets that
 have too many labels and too small cardinalities of subsets for each label
 for conventional methods to improve search accuracy,
 such as natural images, speech data, and fingerprint image data.


As can be easily seen, feature selection by the proposed method can be applied
 to other hashing schemes (such as Spectral hashing)
 than hashing using hyperplanes, which proves its broad versatility.


\bibliographystyle{unsrt}
\bibliography{BibForHashing}

\begin{thebibliography}{10}

\bibitem{Seitai_JRP}
Anil~K. Jain, Arun Ross, and Salil Prabhakar.
\newblock An introduction to biometric recognition.
\newblock {\em IEEE Trans. on Circuits and Systems for Video Technology},
  14:4--20, 2004.

\bibitem{Finger_power}
Haiyun Xu and Raymond N.~J. Veldhuis.
\newblock Spectral minutiae representations for fingerprint recognition.
\newblock In {\em Proceedings of the 2010 Sixth International Conference on
  Intelligent Information Hiding and Multimedia Signal Processing}, IIH-MSP
  '10, pages 341--345, Washington, DC, USA, 2010. IEEE Computer Society.

\bibitem{Seitai_FujitsuPress}
Fujitsu develops world's first personal authentication technology to integrate
  palm vein and fingerprint authentication.
  \url{http://www.fujitsu.com/global/news/pr/archives/month/2011/20110601-01.htm}l.
\newblock Technical report, 2011.

\bibitem{KDTree}
Sunil Arya, David~M. Mount, Nathan~S. Netanyahu, Ruth Silverman, and Angela~Y.
  Wu.
\newblock An optimal algorithm for approximate nearest neighbor searching fixed
  dimensions.
\newblock {\em J. ACM}, 45(6):891--923, nov 1998.

\bibitem{iDistance}
H.~V. Jagadish, Beng~Chin Ooi, Kian-Lee Tan, Cui Yu, and Rui Zhang.
\newblock idistance: An adaptive b+-tree based indexing method for nearest
  neighbor search.
\newblock {\em ACM Trans. Database Syst.}, 30(2):364--397, jun 2005.

\bibitem{LSH_IndykMotwani}
Piotr Indyk and Rajeev Motwani.
\newblock Approximate nearest neighbors: Towards removing the curse of
  dimensionality.
\newblock pages 604--613, 1998.

\bibitem{LSH_RandomProjection}
Moses~S. Charikar.
\newblock Similarity estimation techniques from rounding algorithms.
\newblock In {\em Proceedings of the thiry-fourth annual ACM symposium on
  Theory of computing}, STOC '02, pages 380--388, New York, NY, USA, 2002. ACM.

\bibitem{MLH}
Mohammad~Norouzi and David~J. Fleet.
\newblock Minimal loss hashing for compact binary codes.
\newblock In Lise Getoor and Tobias Scheffer, editors, {\em ICML}, pages
  353--360. Omnipress, 2011.

\bibitem{MNIST}
The {MNIST} database of handwritten digits.
  \url{http://yann.lecun.com/exdb/mnist/}.

\bibitem{LabelMe}
Label{M}e: The open annotation tool. \url{http://labelme.csail.mit.edu/}.

\bibitem{SpeechMFCC}
Steven~B. Davis and Paul Mermelstein.
\newblock Comparison of parametric representations for monosyllabic word
  recognition in continuously spoken sentences.
\newblock {\em IEEE Transactions on Acoustics, Speech, and Signal Processing},
  28(4), 1980.

\bibitem{PCAH}
Xin-Jing Wang, Lei Zhang, Feng Jing, and Wei-Ying Ma.
\newblock Annosearch: Image auto-annotation by search.
\newblock In {\em Proceedings of the 2006 IEEE Computer Society Conference on
  Computer Vision and Pattern Recognition - Volume 2}, CVPR '06, pages
  1483--1490, Washington, DC, USA, 2006. IEEE Computer Society.

\bibitem{FS1}
Yadong Mu, Xiangyu Chen, Tat-Seng Chua, and Shuicheng Yan.
\newblock Learning reconfigurable hashing for diverse semantics.
\newblock In {\em Proceedings of the 1st ACM International Conference on
  Multimedia Retrieval}, ICMR '11, pages 7:1--7:8, New York, NY, USA, 2011.
  ACM.

\bibitem{FS2}
Yu-Gang Jiang, Jun Wang, and Shih-Fu Chang.
\newblock Lost in binarization: query-adaptive ranking for similar image search
  with compact codes.
\newblock In {\em Proceedings of the 1st ACM International Conference on
  Multimedia Retrieval}, ICMR '11, pages 16:1--16:8, New York, NY, USA, 2011.
  ACM.

\bibitem{KLSH}
Brian Kulis and Kristen Grauman.
\newblock Kernelized locality-sensitive hashing for scalable image search.
\newblock In {\em IEEE International Conference on Computer Vision (ICCV},
  2009.

\bibitem{SH}
Yair Weiss, Antonio Torralba, and Robert Fergus.
\newblock Spectral hashing.
\newblock In {\em NIPS}, pages 1753--1760, 2008.

\bibitem{MarginBasedFS}
Amir Navot and Naftali Tishby.
\newblock Margin based feature selection - theory and algorithms.
\newblock In {\em In International Conference on Machine Learning (ICML}, pages
  43--50. ACM Press, 2004.

\bibitem{GIST}
Aude Oliva and Antonio Torralba.
\newblock Modeling the shape of the scene: A holistic representation of the
  spatial envelope.
\newblock {\em Int. J. Comput. Vision}, 42(3):145--175, may 2001.

\bibitem{512dimGISTdata}
Antonio Torralba, Rob Fergus, and Yair Weiss.
\newblock Small codes and large image databases for recognition.
\newblock In {\em In Proceedings of the IEEE Conf on Computer Vision and
  Pattern Recognition}, 2008.

\bibitem{KawasakiCouncil}
Internet relay broadcast of {K}awasaki-city parliament.
  \url{http://www.kawasaki-council.jp/}.

\end{thebibliography}


\end{document}